%% file: main.tex
\documentclass[conference]{IEEEtran}
\IEEEoverridecommandlockouts
\input{math_commands.tex}

\usepackage{hyperref}
\usepackage{url}
\hypersetup{
    colorlinks=true,    
    linkcolor=blue,     
    citecolor=blue,     
    urlcolor=blue       
}

\usepackage{float}  
\usepackage{color}  
\usepackage{adjustbox} 
\usepackage{multirow}
\usepackage{subfigure}
\usepackage{caption}
\usepackage{subcaption}

\usepackage{algorithm}  
\usepackage{algorithmic}

\usepackage{authblk}

\def\BibTeX{{\rm B\kern-.05em{\sc i\kern-.025em b}\kern-.08em
    T\kern-.1667em\lower.7ex\hbox{E}\kern-.125emX}}
\begin{document}

\title{SparseDM: Toward Sparse Efficient Diffusion Models}



\author[1,3]{Kafeng Wang}
\author[1,*]{Jianfei Chen}
\author[1]{He Li}
\author[2]{Zhenpeng Mi}
\author[1,3,*]{Jun Zhu}
\affil[1]{Dept. of Comp. Sci. \& Tech., Institute for AI, BNRist Center\\ 
\protect\\ 
Tsinghua-Bosch Joint ML Center, THBI Lab, Tsinghua University,  Beijing, China} 
\affil[2]{Honor Device Co., Ltd., Beijing, China}
\affil[3]{ShengShu, Beijing, China}
\affil[*]{Corresponding authors: jianfeic@tsinghua.edu.cn \& dcszj@tsinghua.edu.cn}

%


\maketitle

\begin{abstract}
Diffusion models represent a powerful family of generative models widely used for image and video generation.
However, the time-consuming deployment, long inference time, and requirements on large memory hinder their applications on resource constrained devices.
In this paper, we propose a method based on the improved Straight-Through Estimator to improve the deployment efficiency of diffusion models. 
Specifically, we add sparse masks to the Convolution and Linear layers in a pre-trained diffusion model, then transfer learn the sparse model during the fine-tuning stage and turn on the sparse masks during inference.
Experimental results on a Transformer and UNet-based diffusion models demonstrate that our method reduces MACs by $50\%$ while maintaining FID. 
Sparse models are accelerated by approximately 1.2x on the GPU.
Under other MACs conditions, the FID is also lower than 1 compared to other methods.
\end{abstract}


\input{introduction}

\input{related-work}
\input{problem-formulation}
\input{method}
\input{experiment}
\input{conclusion}

\bibliographystyle{IEEEtran}
\bibliography{reference}

\input{appendix}

\end{document}

%% file: math_commands.tex
\usepackage{amsmath,amsfonts,bm}

\def\eqref#1{equation~\ref{#1}}

\def\1{\bm{1}}

\def\vmu{{\bm{\mu}}}

\def\vepsilon{{\bm{\epsilon}}}

\def\vx{{\bm{x}}}
\def\vy{{\bm{y}}}

\DeclareMathAlphabet{\mathsfit}{\encodingdefault}{\sfdefault}{m}{sl}
\SetMathAlphabet{\mathsfit}{bold}{\encodingdefault}{\sfdefault}{bx}{n}

\newcommand{\E}{\mathbb{E}}



%% file: introduction.tex
\section{Introduction}

Diffusion Models (DM) \cite{song2021denoising,karras2022elucidating} have been one of the core generative modules in various computer vision tasks. Generally, they are composed of two parts: a forward/diffusion process that perturbs the data distribution to learn the time-dependent score functions, and a reverse/sampling process that generates data samples from a prior distribution in an iterative manner. Though diffusion models have advantages on both sample quality and mode coverage over other competitors, their slow inference speed and heavy computational load during the inference process inevitably restrict their applications on most resource constrained devices.

To reduce the computational load in the inference process of diffusion models, various methods have been proposed to minimize the number of inference steps, such as the training-free samplers \cite{bao2021analytic,lu2022dpm,zheng2023dpm} and the distillation methods \cite{salimans2022progressive}. However, their sample quality is still unsatisfactory as a few sampling steps cannot faithfully reconstruct the high-dimensional data space, e.g., the image or video samples. 
This problem is even more pronounced on resource constrained devices.
Simultaneously, a few works explore reducing the Multiple-Accumulate operations (MACs) at each inference step \cite{fang2023structural}. However, their work cannot be accelerated on GPUs.
Since DM is an intensive model parameter calculation, NVIDIA Ampere architecture GPU supports a 2:4 sparse (4-weights contain 2 non-zero values) model calculation, which can achieve nearly 2 times calculation acceleration \cite{pool2021channel}.  
Although structural pruning \cite{fang2023structural} has been used in DM, 2:4 structured sparsity inference has not been implemented.
Our method aims to reduce MACs at each step through 2:4 and other scale sparsity.

Recently, several popular computational architectures, such as NVIDIA Ampere architecture and Hopper GPUs, have developed acceleration methods for model inference and have been equipped with fine-grained structured sparse capabilities. A common requirement of these acceleration techniques is the 2:4 sparse mode, which only preserves 2 of the 4 adjacent weights of a pre-trained model, \textit{i.e.}, requires a sparsity rate of $50\%$. Given this sparsity, the acceleration techniques only process the non-zero values in matrix multiplications, theoretically achieving a 2x speedup. In essence, a diffusion model supporting the 2:4 sparsity mode can reduce $50\%$ computational load at each inference step, which is valuable when considering the iterative refinement process in a generation. To the best of our knowledge, previous works have not explored maintaining the sample quality of diffusion models with a 2:4 or other scale sparsity mode, which motivates us to present the techniques in this work.

We use a state-of-the-art Transformer-based DM, named U-ViT \cite{bao2023all}, to analyze the shortcomings of existing sparse pruning tools.
Also, we manually design a half-size DM network to test FID and MACs.
From Table \ref{tab:compare}, we can see that simply reducing the U-ViT model parameters by nearly $50\%$ may cause FID to collapse catastrophically. 
Automatic Sparsity (ASP) \cite{pool2021channel}) for model sparse training consumes a lot of GPU time, but FID is poor.
Therefore, for DM, we need to redesign the pruning method to reduce the amount of calculation and maintain FID as much as possible.
\input{results2.3.table} 
This paper proposes a new method for implementing 2:4 structured sparsity and other scale sparse inference.
(1) Masks with different sparsity rates are applied to convolutional and linear layers.
(2) Sparse regularization is added to back-propagation, improving the STE method to train sparse models.
(3) Knowledge is gradually transferred from dense models to sparse models to improve models' performance with high sparsity rates.

Our contributions can be listed as follows: 
\begin{itemize}
\item We propose a transfer learn sparse masks method that achieves 2:4 structured sparse inference and other scale sparse pruning for DMs with a Transformer or UNet backbones.
\item We conduct experiments on four datasets. The average FID of 2:4 sparse DM is not increased compared with the dense model. The inference acceleration on the GPU is approximately 1.2x.
\item Testing nine scales, with similar MACs for each scale, our sparse inference FID is $1$ lower than other methods.
\end{itemize}

%% file: results2.3.table.tex
\begin{table}[t]
\centering
\caption{The results of manually redesigning the U-ViT model and sparse pruning the U-ViT model using ASP.
UViT Small is from the U-ViT method. Half UViT Small is set to half the depth and number of heads of UViT Small.
ASP is the method of sparse pruning from Nvidia.}
\label{tab:compare}
\begin{tabular}{c|c|c|c|c} 
\hline
Datasets                                                                & Models          & Methods & FID    & MACs (G)  \\ 
\hline
\multirow{3}{*}{\begin{tabular}[c]{@{}c@{}}CIFAR10\\32x32\end{tabular}} & UViT Small      & U-ViT   & 3.97   & 11.34     \\ 
\cline{2-5}
                                                                        & Half UViT Small & U-ViT   & 678.20 & 5.83      \\ 
\cline{2-5}
                                                                        & UViT Small      & ASP     & 319.87 & 5.76      \\ 
\hline
\multirow{3}{*}{\begin{tabular}[c]{@{}c@{}}CelebA\\64x64\end{tabular}}  & UViT Small      & U-ViT   & 3.30   & 11.34     \\ 
\cline{2-5}
                                                                        & Half UViT Small & U-ViT   & 441.37 & 5.83      \\ 
\cline{2-5}
                                                                        & UViT Small      & ASP     & 438.31 & 5.76      \\
\hline
\end{tabular}
\vspace{-0.3cm}
\end{table}

%% file: related-work.tex
\section{Related Work}
A neural network with {$N$:$M$} sparsity ($M$-weights contains $N$ non-zero values) satisfies that, in each group of $M$ consecutive weights of the network, there are at most $N$ weights have non-zero values \cite{zhou2021learning}. 
Compared with dense networks, sparse network training has gradient changes, and methods such as STE \cite{bengio2013estimating} should be used to improve training performance. In the forward stage, sparse weight is obtained by pruning dense weight. In the backward stage, the gradient w.r.t. sparse weight will be directly applied to dense weight.
Sparsity can reduce the memory footprint of DM to fit resource constrained devices and shorten training time for ever-growing networks \cite{hoefler2021sparsity}.
Pruning is one of the most used methods to reduce the calculation time of DNN, including DM \cite{fang2023structural}. 
Pay attention to features and select valuable features for the target dataset \cite{wang2020pay}, which can also reduce computational complexity.
LD-Pruner \cite{castells2024ld} and P-ESD \cite{yang2024pruning} prune stable diffusion for specific tasks.
Although there are many pruning methods for DM, either the sparse model FID is poor or structured 2:4 sparsity is not implemented for general base DM.

%% file: problem-formulation.tex
\section{Problem Formulation}
\subsection{Diffusion Models}
The diffusion model is divided into a forward process and a backward process. 
The forward process is a step-by-step process of adding noise to the original image to generate a noisy image,
generally formalized as a Markov chain process.
The reverse process is to remove noise from a noisy image and restore the original image as much as possible.
Gaussian mode was adopted to approximate the ground truth reverse transition of the Markov chain.
The training process of diffusion models is the process of establishing noise prediction models.
The loss function of DM is minimizing a noise prediction objective.
The forward process is formalized as $q(\vx_{1:T}|\vx_0) = \prod_{t=1}^T q(\vx_t|\vx_{t-1})$,
where $\vx_t$ is input data at $t$.
In the backward process, the mean network of the denoising transition probability is formalized as $\vmu_t^*(\vx_t) = \frac{1}{\sqrt{\alpha_t}} \left(\vx_t - \frac{\beta_t}{\sqrt{1-\overline{\alpha}_t}} \E[\vepsilon|\vx_t]\right)$,
where $\alpha_t$ and $\beta_t$ are the noise schedule at $t$, $\alpha_t+\beta_t=1$. $\overline{\alpha}_t = \prod_{i=1}^t \alpha_i$, and $\vepsilon$ is the standard Gaussian noises added to $\vx_t$.
The DM training is a task of minimizing the noisy prediction errors, expressed as $\min_{\mathcal{W}} \mathcal{L}(\mathcal{W};\mathcal{D})$, 
where $\mathcal{D}$ is dataset, $\mathcal{L}$ is loss function, $\mathcal{W}$ is dense weights.
The DM inference, data generation, is expressed as $\vy_t = \mathcal{F}(\mathcal{W}; \vx_t)$, 
where $\mathcal{F}$ is trained DM, $\vy_t$ is the output of DM inference.

\subsection{Sparse Pruning}
Sparse network computation is an effective method to reduce MACs in deep networks and accelerate computation.
Currently, one-shot sparse training and progressive sparse training are commonly used.

The one-shot training method is easy to use, and the steps are as follows \cite{pool2021channel}:
Train a regular dense model.
Prune the weights on the fully connected and convolutional layers in a 2:4 sparse mode.
Retrain the pruned model.
The one-shot pruning is expressed as $\widetilde{\mathcal{W}} = \mathcal{W} \odot \mathcal{M}$,
where $\widetilde{\mathcal{W}}$ is the sparse weight after the dense weight $\mathcal{W}$ is pruned, recorded as 
$\widetilde{\mathcal{W}} \leftarrow \text{Pruning}(\mathcal{W})$, $\mathcal{M}$ is a 0-1 mask, $\mathcal{M}$ is the ($N$:$M$) sparse mask ($M$-weights contain $N$ non-zero values), $\odot$ represents element-wise multiplication.

One-shot sparse pruning can cover most tasks and achieve speedup without losing accuracy. However, for some challenging tasks that are sensitive to changes in weight values, doing sparse training for all weights at once will result in a large amount of information loss \cite{han2015learning}. 
With the same number of tuning iterations, progressive sparse training can achieve higher model accuracy than one-shot sparse training. 
Suppose there are $k$ masks $[\mathcal{M}_1, \mathcal{M}_2, ..., \mathcal{M}_k]$, the progressive sparse process is formalized as $\widetilde{\mathcal{W}} = \mathcal{W} \odot \mathcal{M}_i ,  \mathcal{M}_i \in [\mathcal{M}_1, \mathcal{M}_2, ..., \mathcal{M}_k]$.
The reason why traditional progressive sparse training methods work well is to reuse the knowledge of dense models as much as possible.
However, this progressive sparse training method has only proven effective when the training data distribution is stable, such as when training a CNN classification model.
\subsection{Distribution Shift on DM Training and Sparse Pruning}
Robust Fairness Regularization (RFR) \cite{jiang2024chasing} improved that distribution shift can be transformed as data perturbation, and data perturbation and model weight perturbation are equivalent for classifier models. Assuming that the perturbation of the label is not considered, the equivalent of data perturbation and model weight perturbation is formalized as:
\begin{equation}
    \mathbb{E}_{\delta_\mathcal{D}(\mathcal{D})}\mathbb{E}_{(\mathcal{D})\sim \mathcal{P}}[\mathcal{L}(\mathcal{F}_{\mathcal{W}}(\mathcal{D}+\delta_{\mathcal{D}}(\mathcal{D}))] = \mathbb{E}_{(\mathcal{D})\sim \mathcal{P}}[\mathcal{L}(\mathcal{F}_{\mathcal{W}+\Delta\mathcal{W}}(\mathcal{D}))],
    \label{equ:equivalent}
\end{equation}
where the training dataset $\mathcal{D}$ with distribution $\mathcal{P}$, suppose the training dataset is perturbed with data perturbation $\delta$, and the neural network is given by $\mathcal{F}_{\mathcal{W}}(\cdot)$, for the general case, there exists model weight pruned as perturbation $\Delta\mathcal{W}$, so that the training loss $\mathcal{L}$ on perturbed training dataset is the same with that for model weight perturbation $\Delta\mathcal{W}$ on training distribution. 

The training process of DMs generally involves only the distribution shift of the noisy data. We believe that DM training adds Gaussian noise to the data, which perturbs the original image data distribution.
Existing sparse pruning methods generally do not consider the distribution shift of noisy data but only the model's weight changes.
Inspired by RFR's conclusions, we convert the distribution shifts caused by changes in model weights into distribution shifts caused by data changes for the DM's training process. 
The DM's sparse ratio is fixed, and then the knowledge of the dense model is transferred to the sparse model.

%% file: method.tex
\section{Sparse Finetuning DM}
\begin{figure*}[!ht]
    \centering
    \includegraphics[scale=0.45]
    {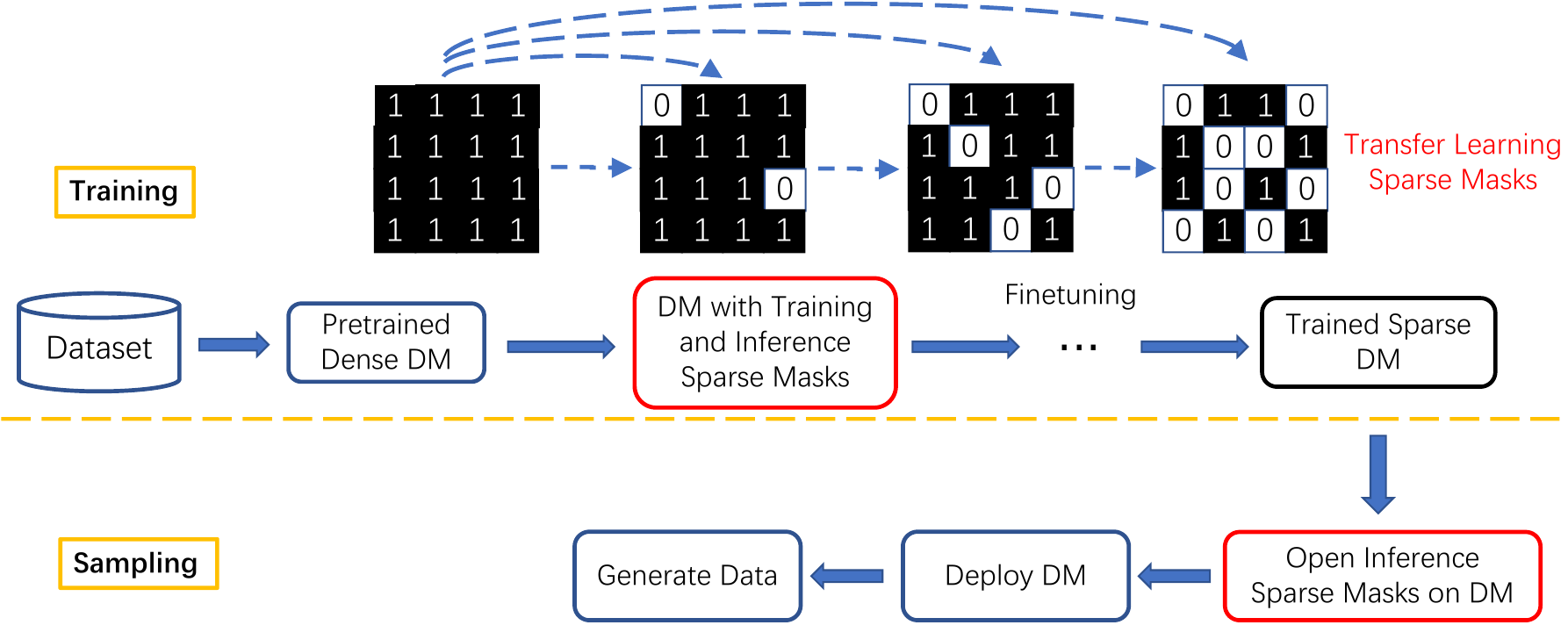}
    \vspace{-0.2cm}
    \caption{Framework Overview. This includes the process of transfer learning sparse models.}
    \vspace{-0.3cm}
    \label{fig:overview}
\end{figure*}

In this section, we introduce our proposed framework in detail. We start with the overall framework of our proposal. Then, we present the finetuning DM with sparse masks to reduce MACs and prepare for 2:4 sparse GPU acceleration. Moreover, we introduce the training and inference with sparse masks to enhance the DM. 
Fig.~\ref{fig:overview} shows an overview of the proposed framework. 
\subsection{Straight-Through Estimator for Sparse Training}
A straightforward solution for training an $N$:$M$ sparsity network is to extend the Straight-through Estimator (STE)~\cite{bengio2013estimating} to perform online magnitude-based pruning and sparse parameter updating. The concrete formula is described as $\mathcal{W}_{t+1} \leftarrow \mathcal{W}_{t} - \gamma_t g(\widetilde{\mathcal{W}^{(N:M)}_t})$,
where $\gamma _t$ is learning rate at time $t$, and $g$ is the gradient.  
In STE, a dense network is maintained during the training process. 
In the forward pass, the dense weights $\mathcal{W}$ are projected into sparse weights
$\widetilde{\mathcal{W}^{(N:M)}}=S(\mathcal{W}, N, M)$ 
satisfying $N$:$M$ sparsity, and here $S(\cdot)$ is a projection function. The sparse DM training task is expressed as $\min_{S(\mathcal{W}, N, M)} \mathcal{L}(\mathcal{W};\mathcal{D})$, 
where $\mathcal{L}$ is the loss function for training DM.
The sparse DM inference is expressed as $\vy_t^{(N:M)} = \mathcal{F}(\widetilde{\mathcal{W}^{(N:M)}}; \vx_t)$, 
where $\vx_t$ is input data at $t$, $\vy_t^{(N:M)}$ is the inference output of $N$:$M$ sparse DM at time $t$.
We incorporate sparse mask information into the backward propagation process to mitigate the negative impact of the approximate gradient calculated by vanilla STE.
The updated formula is modified as follows:
\begin{equation}
\mathcal{W}_{t+1} \leftarrow \mathcal{W}_{t} - \gamma_t \left(g(\widetilde{\mathcal{W}^{(N:M)}_t}) + \lambda_W (\mathcal{W}_{t} - \widetilde{\mathcal{W}^{(N:M)}_t})\right),
  \label{equ:srfinetuning}
\end{equation}
where $\lambda_W$ is a tunable hyperparameter, $\gamma _t$ is learning rate at time $t$, and $g$ is the gradient.

Traditional progressive sparse training works well for highly sparse model training \cite{han2015learning,pool2021channel}. However, its training data distribution does not change.
According to the analysis of the two distribution shifts in Section 2.3, only one distribution shift is usually optimized when training using stochastic gradient descent or its improved algorithms.
The optimization will fail if two distribution shifts are optimized simultaneously, such as switching the sparsity rate when training a DM.
Directly training an extremely sparse model, such as 1:32 ($\frac{1}{32}=0.03125$) sparsity, initialized by dense model weight and using extremely sparse weight gradients for reverse propagation may cause the sparse model to collapse. In this process, knowledge of the dense model is easily lost.

\subsection{Transfer Learn Sparse Diffusion Models}

To train a highly sparse DM from a dense DM, we fix the sparsity rate during DM training and transfer the knowledge to the sparse model via samples generated by the dense model.
This method can accelerate the sparse training of DM and reuse the knowledge of the dense model as much as possible.
This method does not change the number of sampling steps used to generate samples. 
The following is a formal description of progressive knowledge transfer.

$\vx_0$ is the data used to train DM.
During training, a sparse neural network $\mathcal{F}_\mathcal{W}(\vx_t, t)$ is trained to predict the noise in $\vx_t$ w.r.t. $\vx_0$ by minimizing the L2 loss between them. 
The loss is formulated as $\mathcal{L}_\mathrm{diff} := ||\bm{\epsilon}_t - \mathcal{F}_\mathcal{W}(\vx_t, t)||_2^2$.
The overall loss function of sparse training comprises the original dense task loss $\mathcal{L}_\mathrm{dense}$, a diffusion loss $\mathcal{L}_\mathrm{diff}$ that optimizes the diffusion model. It can be formulated as:
\begin{equation} \label{eq:overall_loss}
    \mathcal{L}_\mathrm{sparse}^i = \lambda_{1} \mathcal{L}_\mathrm{dense} + \lambda_{2} \mathcal{L}_\mathrm{diff}, ~~ i \in [1,k],
\end{equation}
where $\lambda_{1}$ and $\lambda_{2}$ are hyper-parameters to balance the losses, with range of $[0,1]$, and $k$ is the sparse mask number. When training a sparse model, especially extremely sparse models, we will select a teacher model, such as a dense or sparse model, and adjust the values of $\lambda_{1}$ and $\lambda_{2}$.

To achieve 2:4 sparse GPU acceleration, the current sparse training method is to directly train from a dense model with a sparsity rate of 0 to a sparse model with a sparsity rate of $50\%$, which can easily cause information loss. Due to the lengthy training process aimed at obtaining $50\%$ of the sparse model, we adopt a progressive sparse training process, ensuring we can obtain $50\% $ of the sparse model with minimal information loss. We add progressive sparse masks to existing STE-based methods. The progressive sparse DM training task is expressed as $\min_{S(\mathcal{W} \odot \mathcal{M})} \mathcal{L}(\mathcal{W};\mathcal{D};[\mathcal{M}_1, \mathcal{M}_2, ..., \mathcal{M}_k])$. 
The projection function $S(\cdot)$, which is non-differentiable during back-propagation, generates the $N$:$M$ sparse sub-network on the fly. To get gradients during back-propagation, STE computes the gradients of the sub-network $g(\widetilde{\mathcal{W}}) = \bigtriangledown_{\widetilde{\mathcal{W}}}   \mathcal{L}(\widetilde{\mathcal{W}};\mathcal{D})$
based on the sparse sub-network $\widetilde{\mathcal{W}}$, 
which can be directly back-projected to the dense network as the approximated gradients of the dense parameters. The approximated parameter update rule for the dense network can be formulated as:
\begin{equation}
\mathcal{W}_{t+1} \leftarrow \mathcal{W}_{t} - \gamma_t g(\widetilde{\mathcal{W}^{(N_i:M_i)}_t}), ~~ i \in [1,k],
  \label{equ:progressive_finetuning}
\end{equation}
where $\gamma _t$ is learning rate at time $t$, $g$ is the gradient, $\widetilde{\mathcal{W}^{(N_i:M_i)}_t}$ is the sparse weight with mask $N_i$:$M_i$ at time $t$.
This STE-based method could be easily improved by sparse mask regularization as follows: 
\begin{equation}
   \mathcal{W}_{t+1} \leftarrow \mathcal{W}_{t} - \gamma_t \left(g(\widetilde{\mathcal{W}^{(N_i:M_i)}_t}) + \lambda_W (\mathcal{W}_t - \widetilde{\mathcal{W}^{(N_i:M_i)}_t})\right),  
  \label{equ:sr-progressive_finetuning}
\end{equation}
where $\lambda_W$ is a tunable hyperparameter, $i \in [1,k]$.
\subsection{2:4 Sparse Mask for Sampling on GPU}
The structured sparse function provides fully connected layers and convolutional layers with 2:4 sparse weights to achieve 2:4 sparse acceleration on the GPU. If their weights are pruned ahead of time, these layers can be accelerated using structured sparse functions on the GPU. 
To test actual inference acceleration on NVIDIA GPUs, such as RTX3090, A40, L40, A100, and H100, we used the 2:4 sparse operator in the acceleration library provided by NVIDIA. 
Those NVIDIA GPUs have implemented CUDA operators to accelerate the multiplication of such matrices.
2:4 sparsity inference is expressed as $\vy^{(2:4)}_t = \mathcal{F}(\widetilde{\mathcal{W}^{(2:4)}}; \vx_t)$, 
where $\vx_t$ is input data at $t$, $\vy_t^{(2:4)}$ is the inference output of $2$:$4$ sparse DM at time $t$.
Transposable masks \cite{hubara2021accelerated} is one of the critical technologies for NVIDIA GPUs to accelerate sparse matrices. 
Transposable masks suggest that a weight matrix and its transpose can be simply pruned by multiplying binary masks, so the backward pass shares the same sparse weight matrix with the forward pass. 

%% file: experiment.tex
\section{Results and Discussion}
In this section, we will compare and discuss the different sparsity rate optimization results and some ablation study results.
\subsection{Experimental Settings}
\textbf{Evaluation Metrics.} 
In this paper, we concentrate primarily on two types of metrics:
The efficiency metric is MACs;
The quality metric is FID. 
We generated 50,000 images and calculated FID together with the original images.

\textbf{Baseline Methods.}
We compare the performance of our method with the following baseline algorithm at 2:4 or other sparsity.
Diff-Pruning \cite{fang2023structural} is a method of structural pruning of diffusion models.
ASP (Automatic SParsity) \cite{pool2021channel} is a tool that enables 2:4 sparse training and inference for PyTorch models provided by Nvidia.
Nvidia developed a simple training workflow that can easily generate a 2:4 structured sparse network that matches the accuracy of the dense network.
STE-based Pruning \cite{bengio2013estimating,zhang2022learning} only uses sparse masks for training and inference.
To compare the performance of other sparse masks, we give the results of inference with different sparsity rates.
Finally, we provide the ablation study results of STE-based for the training process and traditional progressive sparse training.
We give the ablation study results of untrained and STE-trained masks for the inference process.

\subsection{Result Comparison}
\textbf{2:4 Sparsity Results.}
\input{results14.table}  
We experimented with four datasets (CIFAR10, CelebA, MS COCO 2014, and ImageNet) and three resolutions (32$\times$32, 64$\times$64, and 256$\times$256) on Transformer-based and UNet-based DMs. 
We reduced the computational load by approximately $50\%$, and the FID did not increase on average.
The higher the resolution, especially for models of similar sizes, the better the FID effect, such as U-ViT on CIFAR10 32$\times$32 and CelebA 64$\times$64.
This way, our method is more suitable for model acceleration in high-resolution and high-fidelity image generation.

\begin{figure*}[t]
\centering
\subfigure[MS-COCO 256$\times$256]{\includegraphics[width=0.24\linewidth]{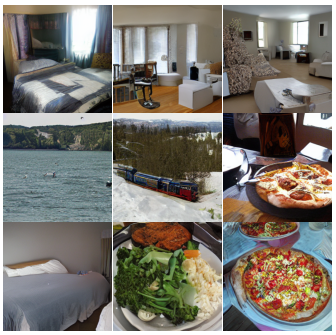}}
\subfigure[ImageNet 256$\times$256]{\includegraphics[width=0.24\linewidth]{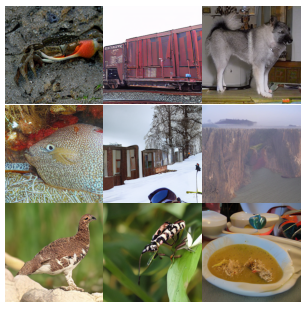}} 
\subfigure[CIFAR10 32$\times$32]{\includegraphics[width=0.24\linewidth]{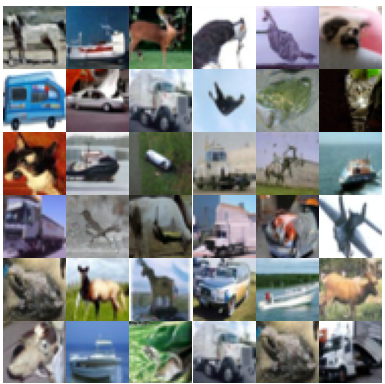}}
\subfigure[CelebA 64$\times$64]{\includegraphics[width=0.24\linewidth]{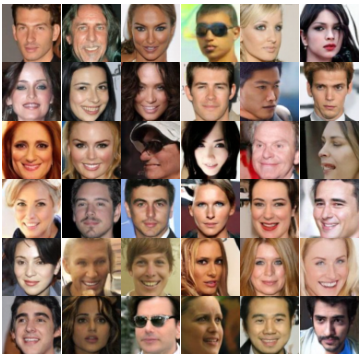}}
\caption{Image generation results of 2:4 sparse U-ViT: selected samples on MS-COCO 256$\times$256, ImageNet 256$\times$256, on CIFAR10 32$\times$32, and CelebA 64$\times$64.}
\vspace{-.3cm}
\label{fig:samples}
\end{figure*}

In addition to changes in FID, it is also important to intuitively evaluate the data generated by the sparse acceleration model.
From the generated images in Fig. \ref{fig:samples}, it can be seen that there is almost no difference between the images generated by our accelerated model and the images generated by the original model due to a slight change in FID.

\input{results_gpu.tex}
Based on the theoretical acceleration results of $50\%$, Table \ref{tab:acceleration_gpu} shows the actual acceleration results of testing the 2:4 sparse operator on the GPU, which is approximately 1.2x.
The acceleration result is defined as the ratio between the running of a dense model and a sparse model in the same setting. The acceleration ratio should almost be the same across all datasets, and the weights have nothing to do with this. 
There are two significant hyper-parameters of the experiment: mlp\_ratio, which sets the dimension of the MLP layer according to the attention layer dimension, and patch\_size, which sets the sequence length for attention computation. 
Experimental experience shows that the sequence length ratio to header dimension affects the speedup ratio.
Some other hyper-parameters are head\_dimension is 1024, num\_head is 8, and depth is 1. 

\textbf{Other Sparsity Results.}
To better understand the performance of our sparse training method, except for the $50\%$ sparsity ratio of 2:4, we have also conducted experiments on other sparsity ratios. Since the ASP method is a sparsity tool provided by Nvidia for GPU hardware acceleration, no other sparsity ratios are provided. The following is mainly a comparison of different sparsity levels with other methods. 

\begin{figure}[!ht]
    \centering
    \vspace{-0.3cm}
    \subfigure[The relationship between FID and mask sparsity ratio.]{\includegraphics[width=0.24\textwidth, height=0.12\textheight]{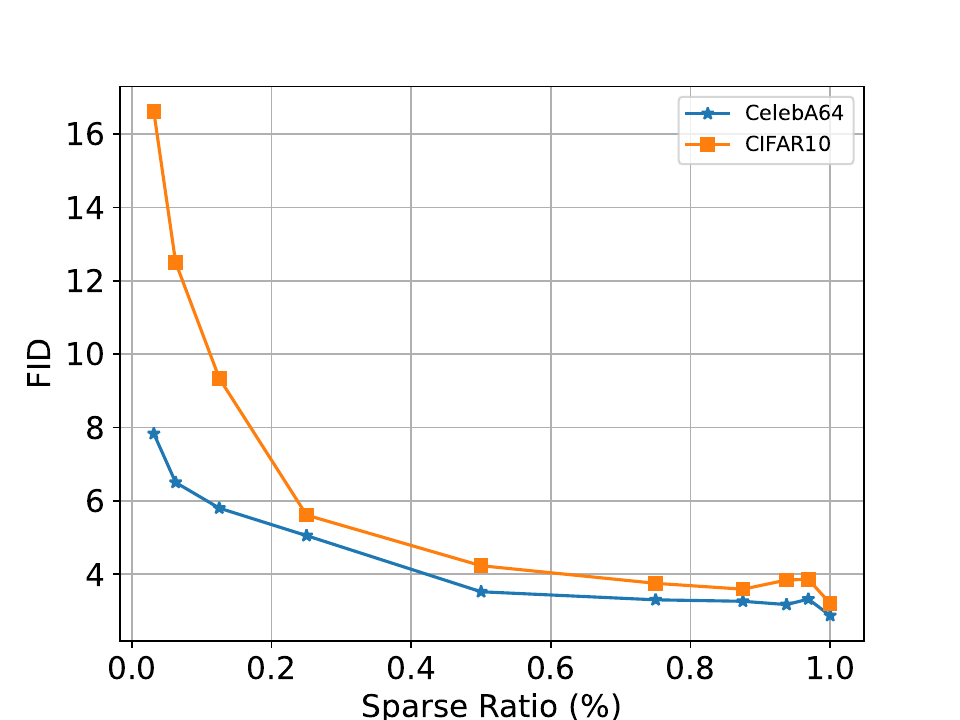}}
    \subfigure[Comparison between traditional progressive and fixed sparse training.]{\includegraphics[width=0.24\textwidth, height=0.12\textheight]{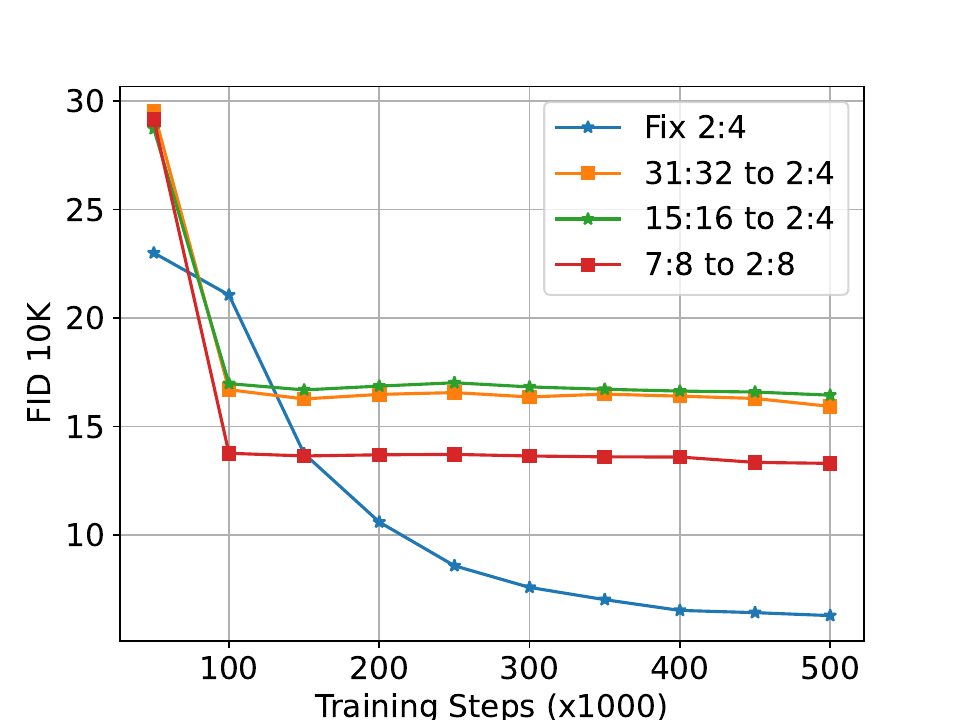}}
    \subfigure[Comparison of FID, MACs trade-off.]{\includegraphics[width=0.24\textwidth, height=0.12\textheight]{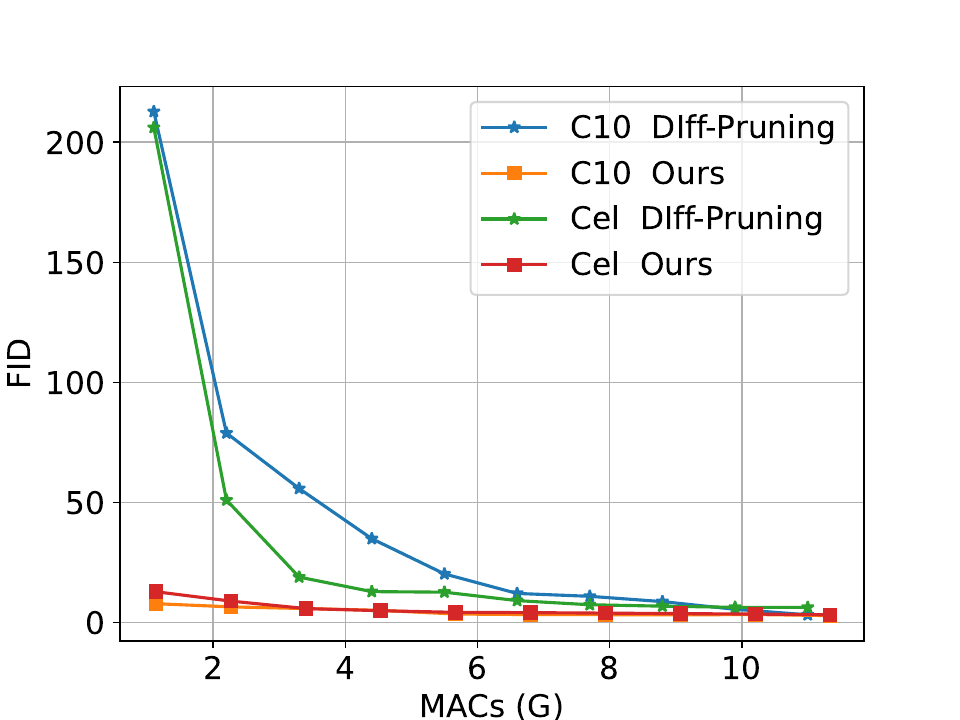}}  
     \subfigure[Comparison of 2:4 sparse training process.]{\includegraphics[width=0.24\textwidth, height=0.12\textheight]{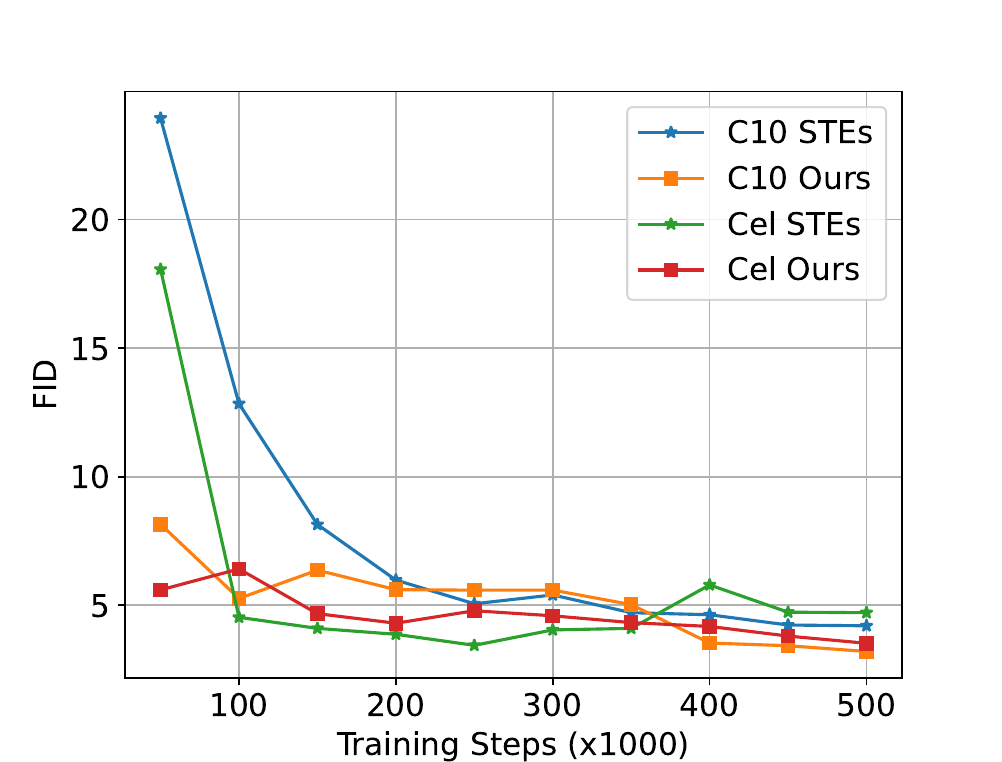}}
    \subfigure[Comparison of FID, MACs trade-off.]{\includegraphics[width=0.24\textwidth, height=0.12\textheight]{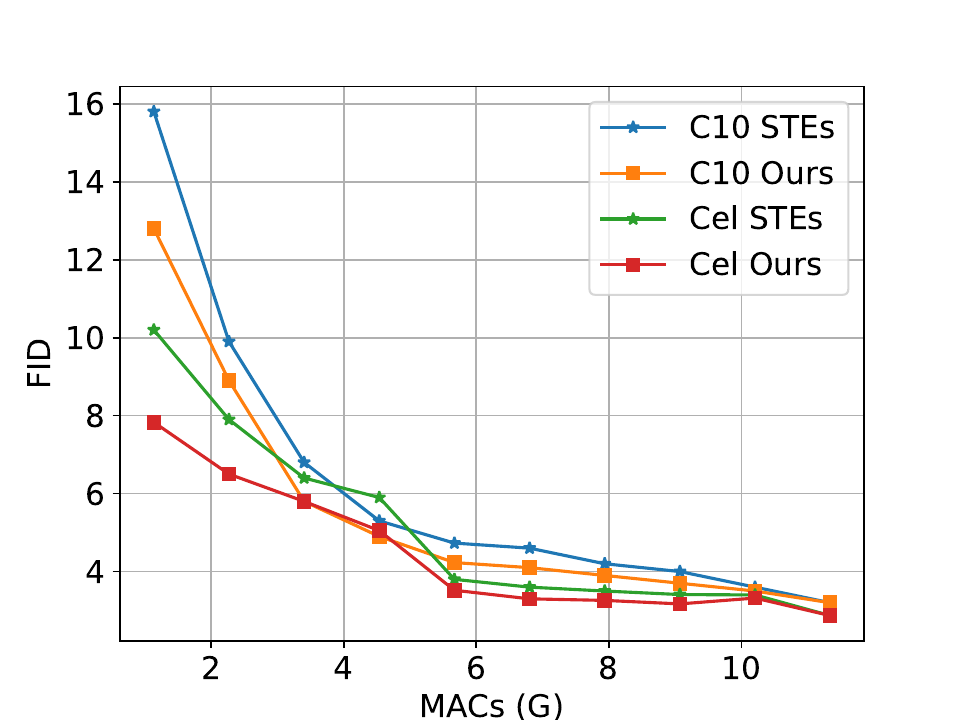}}  
    \subfigure[Comparison between the untrained, STE-trained and our mask.]{\includegraphics[width=0.24\textwidth, height=0.12\textheight]{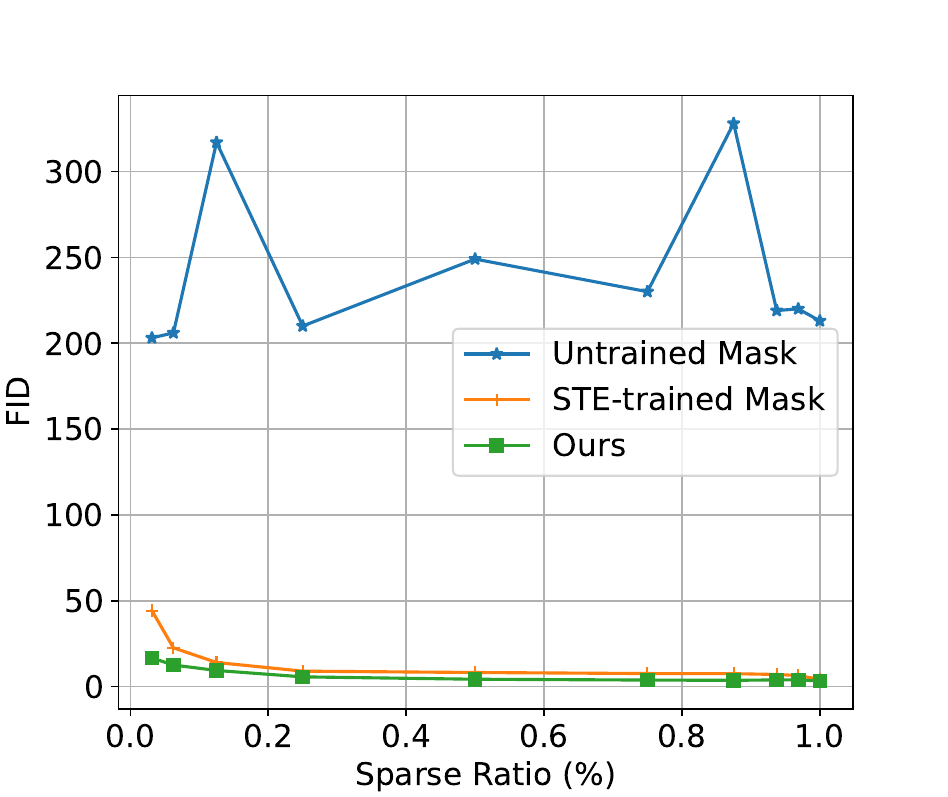}}
    \vspace{-0.2cm}
    \caption{The comparison of sparsity results.}
    \vspace{-0.5cm}
    \label{fig:sparse_inference}
\end{figure}

Fig. \ref{fig:sparse_inference}(a) shows the effect of different sparsity ratios on FID. We evaluated ten sparsity ratios with 32:32, 31:32, 15:16, 7:8, 3:4, 2:4, 1:4, 1:8, 1:16, and 1:32.
As shown in this figure, it does not mean that the greater the sparsity, the better the FID. For example, the sparsity of 31:32 ($\frac{31}{32}=0.96875$) and the sparsity of 15:16 ($\frac{15}{16}=0.9375$), which is higher than the FID of 7:8 with a sparsity of ($\frac{7}{8}=0.875$). In addition to GPU hardware acceleration, the sparsity ratio 2:4 also achieves good FID, proving the 2:4 sparse mask training and inference effectiveness.

Except for the 2:4 sparsity ratio, Fig. \ref{fig:sparse_inference}(c) shows that our method achieves significantly lower FID than the Diff-Pruning method at different sparse ratios on CIFAR10 and CelebA64.
\subsection{Ablation Study}
Our method involves performing sparse mask fine-tuning on the existing trained model. Masks need to be trained during the training process. We designed three ablation experiments to demonstrate better our method design's rationality and the necessity of each step. The first one is the STE-based method. The second one is traditional progressive sparse training of DM. The last one
involves fixing the mask during the training process and not performing sparse training.

\textbf{STE Sparse Training.}
As shown in Fig. \ref{fig:sparse_inference}(d), on datasets CIFAR10 and CelebA64, the learning curve of STE converges significantly slower than our method, mainly because our sparse model is trained with knowledge transfer.
As shown in Fig. \ref{fig:sparse_inference}(e), STE's results are worse than ours, mainly because our method adds sparse mask information during back-propagation. At high compression ratios, such as 1:32, transfer learn sparse masks also play an important role.

\textbf{Traditional Progressive Sparse Training.} As shown in Fig. \ref{fig:sparse_inference}(b), DM is trained on dataset CIFAR10. Training after switching sparse masks for the first time is almost ineffective.
Every 100,000 training steps, the sparsity of masks is increased for progressive sparse training.
The traditional progressive sparse training does not work well on the diffusion model, especially when switching the sparsity rate every time. It is equivalent to training stopped.
However, the fixed sparse mask can be continuously trained until convergence.
This comparison shows that simultaneously transforming the data and model distribution will fail DM training, prompting us to propose a new method for gradually sparsely training DMs.

\textbf{Sparse Mask Inference.} 
Unlike our trained masks, we generate untrained sparse masks for sparse pruning diffusion models. 
Dense model weights are imported into the sparse model, and the sparse model is not trained.
As shown in Fig. \ref{fig:sparse_inference}(f), on dataset CIFAR10, 
the untrained mask's result is the worst at all MACs conditions because the mask did not participate in training. 
The STE-trained mask is worse than ours because the mask did not transfer knowledge from the dense model.
Our method achieves 1 less FID at 9 sparsity rates than the STE-trained masks.

%% file: results14.table.tex
\begin{table}[!ht]
    \centering
\caption{The comparison of 2: 4 ($50\%$ ) sparsity  results. No Pruning is the method of U-ViT, DiT, or DDPM. U-ViT with LDM is U-ViT using Latent Diffusion Models (LDM) \cite{rombach2022high} for data process. Diff-Pruning is a structural pruning method. ASP is the method of sparse pruning from NVIDIA. STE-based Pruning \cite{bengio2013estimating,zhang2022learning} only uses sparse masks for training and inference.
DiT S/2 \cite{peebles2023scalable} is a diffusion model based on Transformer.
DDPM \cite{ho2020denoising} is a diffusion model based on U-Net.
}
\label{tab:quantization_methods}
\begin{tabular}{c|c|c|c|c} 
\hline
Start DM                                                                                           & Datasets                                                                   & Methods                & FID            & MACs (G)        \\ 
\hline
\multirow{10}{*}{\begin{tabular}[c]{@{}c@{}}U-ViT \\(Transformer\\-Based)\end{tabular}}            & \multirow{5}{*}{\begin{tabular}[c]{@{}c@{}}CIFAR10\\32x32\end{tabular}}    & No Pruning             & 3.97           & 11.34           \\ 
\cline{3-5}
                                                                                                   &                                                                            & Diff-Pruning           & 12.63          & 5.32            \\ 
\cline{3-5}
                                                                                                   &                                                                            & ASP                    & 319.87         & 5.76            \\ 
\cline{3-5}
                                                                                                   &                                                                            & STE-based Pruning      & 4.23           & 5.67            \\ 
\cline{3-5}
                                                                                                   &                                                                            & \textbf{Ours}          & \textbf{3.81}  & \textbf{5.67}   \\ 
\cline{2-5}
                                                                                                   & \multirow{5}{*}{\begin{tabular}[c]{@{}c@{}}CelebA\\64x64\end{tabular}}     & No Pruning             & 3.30           & 11.34           \\ 
\cline{3-5}
                                                                                                   &                                                                            & Diff-Pruning           & 11.35          & 5.32            \\ 
\cline{3-5}
                                                                                                   &                                                                            & ASP                    & 438.31         & 5.76            \\ 
\cline{3-5}
                                                                                                   &                                                                            & STE-based Pruning      & 3.75           & 5.67            \\ 
\cline{3-5}
                                                                                                   &                                                                            & \textbf{\textbf{Ours}} & \textbf{3.52}  & \textbf{5.67}   \\ 
\hline
\multirow{10}{*}{\begin{tabular}[c]{@{}c@{}}U-ViT \\(Transformer\\-Based) \\with LDM\end{tabular}} & \multirow{5}{*}{\begin{tabular}[c]{@{}c@{}}MS-COCO\\256x256\end{tabular}}  & No Pruning             & 5.95           & 11.34           \\ 
\cline{3-5}
                                                                                                   &                                                                            & Diff-Pruning           & 15.20          & 5.43            \\ 
\cline{3-5}
                                                                                                   &                                                                            & ASP                    & 350.87         & 5.79            \\ 
\cline{3-5}
                                                                                                   &                                                                            & STE-based Pruning      & 8.14           & 5.68            \\ 
\cline{3-5}
                                                                                                   &                                                                            & \textbf{Ours}          & \textbf{7.09}  & \textbf{5.68}   \\ 
\cline{2-5}
                                                                                                   & \multirow{5}{*}{\begin{tabular}[c]{@{}c@{}}ImageNet\\256x256\end{tabular}} & No Pruning             & 3.81           & 76.66           \\ 
\cline{3-5}
                                                                                                   &                                                                            & Diff-Pruning           & 14.28          & 34.06           \\ 
\cline{3-5}
                                                                                                   &                                                                            & ASP                    & 367.41         & 37.93           \\ 
\cline{3-5}
                                                                                                   &                                                                            & STE-based Pruning      & 5.83           & 36.84           \\ 
\cline{3-5}
                                                                                                   &                                                                            & \textbf{Ours}          & \textbf{5.25}  & \textbf{36.84}  \\ 
\hline
\multirow{5}{*}{\begin{tabular}[c]{@{}c@{}}DiT S/2\\(Transformer\\-Based)\end{tabular}}            & \multirow{5}{*}{\begin{tabular}[c]{@{}c@{}}ImageNet\\256x256\end{tabular}} & No Pruning             & 55.10          & 3.03            \\ 
\cline{3-5}
                                                                                                   &                                                                            & Diff-Pruning           & 98.64          & 1.49            \\ 
\cline{3-5}
                                                                                                   &                                                                            & ASP                    & 308.21         & 1.47            \\ 
\cline{3-5}
                                                                                                   &                                                                            & STE-based Pruning      & 66.48          & 1.51            \\ 
\cline{3-5}
                                                                                                   &                                                                            & \textbf{\textbf{Ours}} & \textbf{51.81} & \textbf{1.51}   \\ 
\hline
\multirow{10}{*}{\begin{tabular}[c]{@{}c@{}}DDPM \\(UNet\\-Based)\end{tabular}}                    & \multirow{5}{*}{\begin{tabular}[c]{@{}c@{}}CIFAR10\\32x32\end{tabular}}    & No Pruning             & 3.23           & 9.44            \\ 
\cline{3-5}
                                                                                                   &                                                                            & Diff-Pruning           & 12.31          & 4.75            \\ 
\cline{3-5}
                                                                                                   &                                                                            & ASP                    & 328.47         & 4.91            \\ 
\cline{3-5}
                                                                                                   &                                                                            & STE-based Pruning      & 3.83           & 4.87            \\ 
\cline{3-5}
                                                                                                   &                                                                            & \textbf{\textbf{Ours}} & \textbf{3.12}  & \textbf{4.87}   \\ 
\cline{2-5}
                                                                                                   & \multirow{5}{*}{\begin{tabular}[c]{@{}c@{}}CelebA\\64x64\end{tabular}}     & No Pruning             & 2.94           & 9.44            \\ 
\cline{3-5}
                                                                                                   &                                                                            & Diff-Pruning           & 11.19          & 4.75            \\ 
\cline{3-5}
                                                                                                   &                                                                            & ASP                    & 441.52         & 4.91            \\ 
\cline{3-5}
                                                                                                   &                                                                            & STE-based Pruning      & 3.66           & 4.87            \\ 
\cline{3-5}
                                                                                                   &                                                                            & \textbf{\textbf{Ours}} & \textbf{3.04}  & \textbf{4.87}   \\
\hline
\end{tabular}
\end{table}

%% file: results_gpu.tex
\begin{table}[!t]
    \centering
    \vspace{-0.2cm}
\caption{
Speedup results for 2:4 ($50\%$) sparse model sampling on CIFAR10 32x32 data tested on 4 A40 GPUs.
Patch\_size and mlp\_ratio are tunable hyperparameters.
}
\label{tab:acceleration_gpu}
\begin{tabular}{c|c|c|c} 
\hline
Acceleration       & patch\_size=1    & patch\_size=2       & patch\_size=4 \\ 
\hline
mlp\_ratio=1         & 1.02 & 1.01    & 1.01               \\ 
\hline
mlp\_ratio=2         & 1.04 & 0.97    & 1.01               \\ 
\hline
mlp\_ratio=4         & \textbf{1.23} & 1.10     & 0.98               \\ 
\hline
mlp\_ratio=8         & \textbf{1.22} & 1.17    & 1.08               \\ 
\hline
\end{tabular}
\vspace{-0.2cm}
\vspace{-0.2cm}
\end{table}

%% file: conclusion.tex
\section{Conclusion}
In this paper, we studied how to improve the efficiency of DM by sparse matrix for 2:4 sparse acceleration processor.  
The existing STE-based methods make it challenging to optimize sparse DM. 
To address this issue, we improve the STE method and propose to gradually transfer knowledge from dense models to sparse models.
Our method is tested on the latest Transformer-based DM, U-ViT, and UNet-based DM, DDPM. 
We trained the 2:4 and other sparse models to perform better than other methods.
Our approach also provides an effective solution for deploying DM on processors that support sparse matrix operators, achieving about 1.2x acceleration.

%% file: appendix.tex
\section{appendix}

\subsection{Parameter counts for each layer of the DM before and after sparse pruning}

We add a 2:4 sparse mask to each convolutional and fully connected layer, so that all models have 50\% of the parameters of each convolutional and fully connected layer. 
For example, the 5th fully connected layer of U-ViT has 600,000 parameters, which becomes 300,000 parameters after sparse masking. We have added this explanation in the revised paper.
\begin{figure}[ht]
    \centering
    \includegraphics[width=0.59\linewidth]{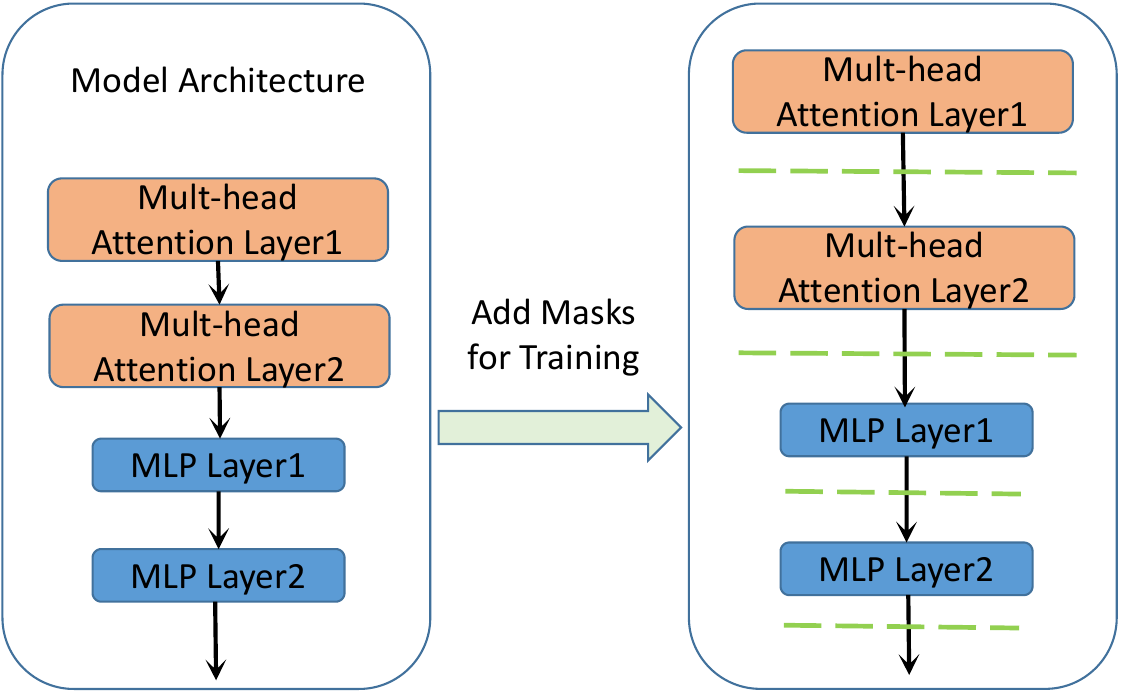}
    \caption{Add sparse mask to each layer}
    \label{fig:dense_sparse_gpu}
\end{figure}

\subsection{The relationship between diffusion training and sparse training}

As shown in Fig. \ref{fig:sparse_inference}(b) of this paper, from the empirical experimental results, it is observed that fixed sparsity applies a consistent distribution shift for all noise levels in diffusion training, while progressive sparsity training gradually shifts the predefined noise level, which may hinder the diffusion training process.

In theory, the relationship between diffusion training and sparse training is mainly explained from the perspective of the difficulty of convergence of the SGD optimizer. Existing optimizers are designed for diffusion training and sparse training, respectively, and the design of each optimizer is challenging. SGD takes the current optimal gradient direction each time it descends, so when using stochastic gradient descent training, usually only one distribution shift is optimized. However, if there are two distribution shifts, the SGD gradient descent direction may not be the current optimal one. Therefore, if two distribution shifts are optimized at the same time, such as switching the sparsity rate when training DM, the optimization will fail.

\subsection{Dense and sparse matrix on GPU}

As shown in Fig. \ref{fig:dense_sparse_gpu}: Dense and sparse matrices on the GPU, in order to implement the sparse structure, Nidia CUDA defines an additional 2-bit indices matrix for calculation. Therefore, when the sparse matrix is not large enough, this additional indices matrix overhead will make the overall result slower.
Therefore, on the GPU, the larger the network model, the better the acceleration results may be.
Fig. \ref{fig:dense_sparse_gpu} is from Google Image. 
https://images.app.goo.gl/7CDgZVcuUYG8rzyc6 
\begin{figure}[ht]
    \centering
    \includegraphics[width=0.99\linewidth]{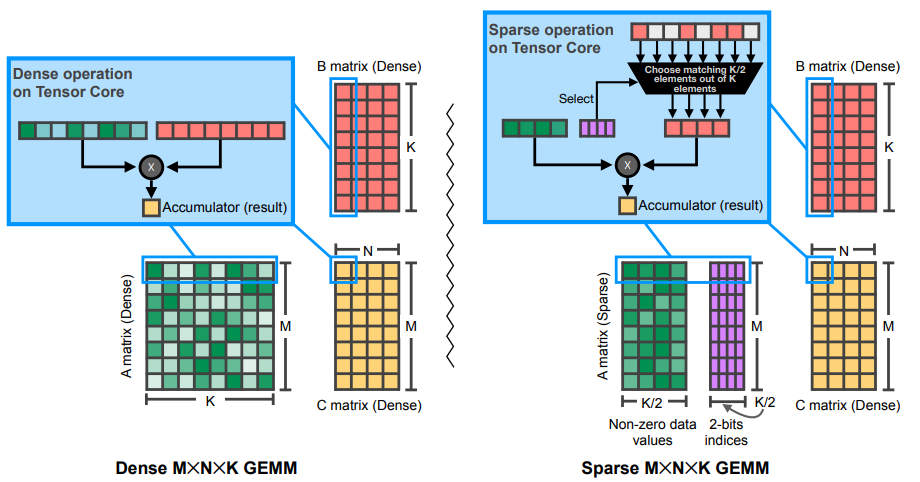}
    \caption{Dense and sparse matrix on GPU.}
    \label{fig:dense_sparse_gpu}
\end{figure}

The Nvidia GPUs that support the 2:4 sparse acceleration operator include RTX3090, A40, L40, A100, H100, etc., which are common computing cards on the market.

\subsection{Limitations and Future Work}

The MACs of the sparse diffusion model are significantly reduced, but the acceleration effect on real processors is limited, especially the 2:4 sparse acceleration that can currently only reach about 1.2 times.

Future work will design sparse acceleration schemes for different processors to achieve better acceleration.
Certain data augmentation \cite{xiong2019sphmc,wang2021sampling} techniques may be used to improve the training of diffusion models.
Certain feature generation \cite{wang2023toward} techniques may be used to improve diffusion model training.
Efficient diffusion models may also be helpful in image processing \cite{yan2024contour,yan2025image}, point cloud processing \cite{gu2022learning}, magnetic sensing signal \cite{wang2021sensemag}, and graph \cite{zhang2025heterogeneous} techniques.

%% file: main.bbl
\begin{thebibliography}{10}
\providecommand{\url}[1]{#1}
\csname url@samestyle\endcsname
\providecommand{\newblock}{\relax}
\providecommand{\bibinfo}[2]{#2}
\providecommand{\BIBentrySTDinterwordspacing}{\spaceskip=0pt\relax}
\providecommand{\BIBentryALTinterwordstretchfactor}{4}
\providecommand{\BIBentryALTinterwordspacing}{\spaceskip=\fontdimen2\font plus
\BIBentryALTinterwordstretchfactor\fontdimen3\font minus \fontdimen4\font\relax}
\providecommand{\BIBforeignlanguage}[2]{{%
\expandafter\ifx\csname l@#1\endcsname\relax
\typeout{** WARNING: IEEEtran.bst: No hyphenation pattern has been}%
\typeout{** loaded for the language `#1'. Using the pattern for}%
\typeout{** the default language instead.}%
\else
\language=\csname l@#1\endcsname
\fi
#2}}
\providecommand{\BIBdecl}{\relax}
\BIBdecl

\bibitem{song2021denoising}
J.~Song, C.~Meng, and S.~Ermon, ``Denoising diffusion implicit models,'' in \emph{International Conference on Learning Representations}, 2021.

\bibitem{karras2022elucidating}
T.~Karras, M.~Aittala, T.~Aila, and S.~Laine, ``Elucidating the design space of diffusion-based generative models,'' \emph{Advances in Neural Information Processing Systems}, vol.~35, pp. 26\,565--26\,577, 2022.

\bibitem{bao2021analytic}
F.~Bao, C.~Li, J.~Zhu, and B.~Zhang, ``Analytic-dpm: an analytic estimate of the optimal reverse variance in diffusion probabilistic models,'' in \emph{International Conference on Learning Representations}, 2022.

\bibitem{lu2022dpm}
C.~Lu, Y.~Zhou, F.~Bao, J.~Chen, C.~Li, and J.~Zhu, ``Dpm-solver: A fast ode solver for diffusion probabilistic model sampling in around 10 steps,'' \emph{Advances in Neural Information Processing Systems}, vol.~35, pp. 5775--5787, 2022.

\bibitem{zheng2023dpm}
K.~Zheng, C.~Lu, J.~Chen, and J.~Zhu, ``Dpm-solver-v3: Improved diffusion ode solver with empirical model statistics,'' in \emph{Thirty-seventh Conference on Neural Information Processing Systems}, 2023.

\bibitem{salimans2022progressive}
T.~Salimans and J.~Ho, ``Progressive distillation for fast sampling of diffusion models,'' in \emph{International Conference on Learning Representations}, 2022.

\bibitem{fang2023structural}
G.~Fang, X.~Ma, and X.~Wang, ``Structural pruning for diffusion models,'' \emph{NeurIPS}, 2023.

\bibitem{pool2021channel}
J.~Pool and C.~Yu, ``Channel permutations for n: M sparsity,'' \emph{Advances in neural information processing systems}, vol.~34, pp. 13\,316--13\,327, 2021.

\bibitem{bao2023all}
F.~Bao, S.~Nie, K.~Xue, Y.~Cao, C.~Li, H.~Su, and J.~Zhu, ``All are worth words: A vit backbone for diffusion models,'' in \emph{Proceedings of the IEEE/CVF Conference on Computer Vision and Pattern Recognition}, 2023, pp. 22\,669--22\,679.

\bibitem{zhou2021learning}
A.~Zhou, Y.~Ma, J.~Zhu, J.~Liu, Z.~Zhang, K.~Yuan, W.~Sun, and H.~Li, ``Learning n: M fine-grained structured sparse neural networks from scratch,'' in \emph{International Conference on Learning Representations}, 2021.

\bibitem{bengio2013estimating}
Y.~Bengio, N.~L{\'e}onard, and A.~Courville, ``Estimating or propagating gradients through stochastic neurons for conditional computation,'' \emph{arXiv preprint arXiv:1308.3432}, 2013.

\bibitem{hoefler2021sparsity}
T.~Hoefler, D.~Alistarh, T.~Ben-Nun, N.~Dryden, and A.~Peste, ``Sparsity in deep learning: Pruning and growth for efficient inference and training in neural networks,'' \emph{The Journal of Machine Learning Research}, vol.~22, no.~1, pp. 10\,882--11\,005, 2021.

\bibitem{wang2020pay}
K.~Wang, X.~Gao, Y.~Zhao, X.~Li, D.~Dou, and C.-Z. Xu, ``Pay attention to features, transfer learn faster cnns,'' in \emph{International conference on learning representations}, 2020.

\bibitem{castells2024ld}
T.~Castells, H.-K. Song, B.-K. Kim, and S.~Choi, ``Ld-pruner: Efficient pruning of latent diffusion models using task-agnostic insights,'' in \emph{Proceedings of the IEEE/CVF Conference on Computer Vision and Pattern Recognition Workshop}, 2024, pp. 821--830.

\bibitem{yang2024pruning}
T.~Yang, J.~Cao, and C.~Xu, ``Pruning for robust concept erasing in diffusion models,'' \emph{arXiv preprint arXiv:2405.16534}, 2024.

\bibitem{han2015learning}
S.~Han, J.~Pool, J.~Tran, and W.~Dally, ``Learning both weights and connections for efficient neural network,'' \emph{Advances in neural information processing systems}, vol.~28, 2015.

\bibitem{jiang2024chasing}
Z.~S. Jiang, X.~Han, H.~Jin, G.~Wang, R.~Chen, N.~Zou, and X.~Hu, ``Chasing fairness under distribution shift: A model weight perturbation approach,'' \emph{Advances in Neural Information Processing Systems}, vol.~36, 2024.

\bibitem{hubara2021accelerated}
I.~Hubara, B.~Chmiel, M.~Island, R.~Banner, J.~Naor, and D.~Soudry, ``Accelerated sparse neural training: A provable and efficient method to find n: m transposable masks,'' \emph{Advances in neural information processing systems}, vol.~34, pp. 21\,099--21\,111, 2021.

\bibitem{zhang2022learning}
Y.~Zhang, M.~Lin, Z.~Lin, Y.~Luo, K.~Li, F.~Chao, Y.~Wu, and R.~Ji, ``Learning best combination for efficient n: M sparsity,'' \emph{Advances in Neural Information Processing Systems}, vol.~35, pp. 941--953, 2022.

\bibitem{rombach2022high}
R.~Rombach, A.~Blattmann, D.~Lorenz, P.~Esser, and B.~Ommer, ``High-resolution image synthesis with latent diffusion models,'' in \emph{Proceedings of the IEEE/CVF conference on computer vision and pattern recognition}, 2022, pp. 10\,684--10\,695.

\bibitem{peebles2023scalable}
W.~Peebles and S.~Xie, ``Scalable diffusion models with transformers,'' in \emph{Proceedings of the IEEE/CVF International Conference on Computer Vision}, 2023, pp. 4195--4205.

\bibitem{ho2020denoising}
J.~Ho, A.~Jain, and P.~Abbeel, ``Denoising diffusion probabilistic models,'' \emph{Advances in neural information processing systems}, vol.~33, pp. 6840--6851, 2020.

\bibitem{xiong2019sphmc}
H.~Xiong, K.~Wang, J.~Bian, Z.~Zhu, C.-Z. Xu, Z.~Guo, and J.~Huan, ``Sphmc: Spectral hamiltonian monte carlo,'' in \emph{Proceedings of the AAAI Conference on Artificial Intelligence}, vol.~33, no.~01, 2019, pp. 5516--5524.

\bibitem{wang2021sampling}
K.~Wang, H.~Xiong, J.~Bian, Z.~Zhu, Q.~Gao, Z.~Guo, C.-Z. Xu, J.~Huan, and D.~Dou, ``Sampling sparse representations with randomized measurement langevin dynamics,'' \emph{ACM Transactions on Knowledge Discovery from Data (TKDD)}, vol.~15, no.~2, pp. 1--21, 2021.

\bibitem{wang2023toward}
K.~Wang, P.~Wang, and C.~Xu, ``Toward efficient automated feature engineering,'' in \emph{2023 IEEE 39th International Conference on Data Engineering (ICDE)}.\hskip 1em plus 0.5em minus 0.4em\relax IEEE, 2023, pp. 1625--1637.

\bibitem{yan2024contour}
L.~Yan, X.~Zhang, K.~Wang, and D.~Zhang, ``Contour-enhanced visual state-space model for remote sensing image classification,'' \emph{IEEE Transactions on Geoscience and Remote Sensing}, 2024.

\bibitem{yan2025image}
L.-y. Yan, X.~Zhang, K.~Wang, S.~Xiong, and D.-j. Zhang, ``Image segmentation refinement based on region expansion and minor contour adjustments,'' \emph{IET Image Processing}, vol.~19, no.~1, p. e70017, 2025.

\bibitem{gu2022learning}
Y.~Gu, H.~Cheng, K.~Wang, D.~Dou, C.~Xu, and H.~Kong, ``Learning moving-object tracking with fmcw lidar,'' in \emph{2022 IEEE/RSJ International Conference on Intelligent Robots and Systems (IROS)}.\hskip 1em plus 0.5em minus 0.4em\relax IEEE, 2022, pp. 3747--3753.

\bibitem{wang2021sensemag}
K.~Wang, H.~Xiong, J.~Zhang, H.~Chen, D.~Dou, and C.-Z. Xu, ``Sensemag: Enabling low-cost traffic monitoring using noninvasive magnetic sensing,'' \emph{IEEE Internet of Things Journal}, vol.~8, no.~22, pp. 16\,666--16\,679, 2021.

\bibitem{zhang2025heterogeneous}
B.~Zhang, H.~Xu, R.~Shuang, and K.~Wang, ``Heterogeneous information-based self-supervised graph learning for recommendation,'' \emph{The Journal of Supercomputing}, vol.~81, no.~4, p. 507, 2025.

\end{thebibliography}
